\documentclass[10pt]{article} 
\usepackage[preprint]{tmlr}

\usepackage{amsmath,amsfonts,bm}

\def\eqref#1{equation~\ref{#1}}

\def\1{\bm{1}}

\DeclareMathAlphabet{\mathsfit}{\encodingdefault}{\sfdefault}{m}{sl}
\SetMathAlphabet{\mathsfit}{bold}{\encodingdefault}{\sfdefault}{bx}{n}

\usepackage{hyperref}
\usepackage{url}
\usepackage{amssymb}
\usepackage{amsmath}
\usepackage{amsthm}
\usepackage{comment}
\usepackage{booktabs}       
\usepackage{amsfonts}       

\usepackage{mathtools}
\usepackage{algorithm}
\usepackage{multirow}

\usepackage{hyperref}
\usepackage{url}
\usepackage{amsthm}
\usepackage{placeins}

\newcommand{\MMD}{\operatorname{MMD}}
\newcommand{\CORAL}{\operatorname{CORAL}}
\usepackage{algorithm}
\usepackage{algpseudocode}
\usepackage{graphicx}
\usepackage{caption}
\usepackage{subcaption}
\usepackage{booktabs}       

\title{Variance-reduced Domain Adaptation using Paired Sampling}

\author{\name Andrea Napoli \\
      \addr Department of Electronics and Computer Science\\
  University of Southampton, UK
}

\def\month{05}  
\def\year{2026} 

\begin{document}

\maketitle

\begin{abstract}
Correlation alignment and the maximum mean discrepancy are two widely used distribution-matching frameworks for unsupervised domain adaptation (UDA). However, high variance in these losses has been shown to undermine their effectiveness in minibatch optimisation settings. Furthermore, the losses lack finite-sum structure, which renders them incompatible with classical stochastic variance reduction (SVR) methods. This paper proposes Paired Sampling for Domain Adaptation (PSDA), a novel SVR technique tailored to such objectives. PSDA pairs observations both within and across domains, to form quadruplets that are always sampled together during training. The pairings are designed to minimise expected gradient variance, and reduce to solving a set of linear assignment problems. Our simulations demonstrate reduced variance compared to related methods, and experiments on three domain shift datasets show improved target domain accuracy.
\end{abstract}

\section{Introduction}

Domain shift remains a key challenge when deploying machine learning models to the real world \citep{Gulrajani2021InGeneralization,Koh2021WILDS:Shifts}. Unsupervised domain adaptation (UDA) aims to address this by minimising a “domain discrepancy” loss during training, which characterises the mismatch between the source and target feature distributions. Two widely used objectives are the correlation alignment (CORAL) loss, which matches covariance matrices \citep{Sun2016DeepAdaptation}, and the maximum mean discrepancy (MMD), which aligns kernel mean embeddings \citep{Tzeng2014DeepInvariance,pmlr-v37-long15,Li2018DomainLearning}.

Although theoretically well-grounded \citep{Ben-David2006AnalysisAdaptation,Ben-David2010ADomains,Redko2022AGuarantees}, estimating these discrepancies in practice is extremely noisy, especially in minibatch optimisation settings. This high variance can destabilise training, and these methods have frequently been observed to perform worse than with no domain alignment at all \citep{Dubey2021AdaptiveGeneralization,Gao2023Out-of-DistributionAugmentations,Gulrajani2021InGeneralization,Koh2021WILDS:Shifts,Napoli2023UnsupervisedCalls,Napoli2024ImprovingSampling,Wang2019CharacterizingTransfer}. Consequentially, a recent paper found that stochastic variance reduction (SVR) can function as an effective stabiliser for these methods, resulting in significant gains in target domain performance \citep{Napoli2026VarianceSampling}. However, most classical SVR methods require the losses to possess finite-sum structure, which the MMD and CORAL terms lack.

Thus, a line of research has emerged developing SVR techniques which are compatible with, or even specialised for, the kinds of non-additive losses found in UDA. The first work to do so proposed diverse sampling of minibatches \citep{Napoli2024ImprovingSampling}; this also helps to balance the data \citep{Napoli2024Diversity-BasedAdaptation}. Outside of UDA, diverse sampling has also previously been found to be beneficial to other nonadditive objectives, such as contrastive losses \citep{Wu2023ContrastiveSamples,Ochieng2025DiversityMagnitudes}, as well as minibatch optimisation more broadly \citep{Zhang2017DeterminantalDiversification,Zhang2019ActiveProcesses,Bardenet2021DeterminantalSGD,Jiang2024DOS:Detection}. However, as well as being computationally rather slow, the effectiveness of this approach is limited by the relatively weak connection between diversity and variance.

The first dedicated SVR technique for UDA was VaRDASS \citep{Napoli2026VarianceSampling}, which leveraged stratified sampling. VaRDASS forms the strata using discrepancy-specific clustering objectives, which directly minimises variance for the MMD and CORAL (with respect to the strata). The computational cost is also negligible once the strata are constructed. Stratified sampling has also been previously explored for minibatch optimisation outside of UDA \citep{Zhao2014AcceleratingSampling,Liu2020AcceleratingStrata,Fu2017CPSG-MCMC:MCMC,Lu2021VarianceModels}.

Diverse and stratified sampling both reduce variance by inducing negative correlation between observations within domains. However, the variance can be reduced further by inducing positive correlation across domains, as well as dependencies across training steps.
To achieve this, \citet{Napoli2026OrderData} proposed ORDERED, which directly optimises variance with respect to the training data sampling order. ORDERED starts with a random data sequence and then iteratively exchanges datapoints between minibatches such that the variance is reduced. This is a type of deterministic minibatch sequencing, which also includes techniques such as curriculum learning \citep{Bengio2009CurriculumLearning} and anti-clustering \citep{Papenberg2021UsingParts,Baumann2026AAlgorithm}. However, the improvement in variance reduction comes at significant computational cost. Additionally, the optimisation procedure relies on a greedy heuristic which is sensitive to local minima, so the solution quality also retains room for improvement.

Most recently, \citet{Napoli2026OnlineData} proposed ARROW, an online algorithm which works on streaming data. ARROW maintains moving averages of the adaptation statistics, and aligns the minibatch statistics with these moving averages using weighted losses. This entails solving a weight-optimisation problem at every timestep, but does not require periodic feature extraction and is suitable for very large datasets.

This paper proposes Paired Sampling for Domain Adaptation (PSDA), an alternative approach to SVR which aims to combine the best aspects of preceding methods, while avoiding each of their drawbacks. PSDA constructs dependencies between sampled observations so that minibatches are more informative for estimating the discrepancy gradient. Specifically, the pairings minimise the expected gradient error, conditioned on the co-occurrence of those pairs in the sample. This extends the idea of antithetic sampling \citep{Liu2018AcceleratingSampling} to the MMD and CORAL losses (note that the formulation provided by \citet{Liu2018AcceleratingSampling} is valid only for classification losses, and is thus also incompatible with non-additive objectives).

To fully exploit the multi-domain setup of UDA, PSDA computes two sets of pairings. The first matches observations across the source and target domains. The second then matches pairs of these matchings, resulting in quadruplets of observations (two from each domain) which are always seen together during training.

A central advantage of this formulation is that each matching reduces to a linear assignment problem, for which efficient and accurate solvers are available. Similarly to VaRDASS, the computational cost of sampling is negligible once the matchings are obtained. However, unlike preceding methods, the complexity of matching is also independent of the minibatch size.

By simply replacing the sampling procedure, PSDA avoids having to alter models, losses, or training algorithms, making it broadly compatible with existing pipelines. It also avoids having to trade off bias (as in shrinkage or momentum), but can still be applied together with these or other SVR methods.

In experiments and simulations, we assess PSDA in terms of speed, degree of variance reduction achieved, and target domain accuracy, finding that our method performs strongly on all three criteria.

\section{Method}\label{method}

\subsection{Preliminaries}\label{preliminaries}

Let
\(\mathcal{D}_{s} = \left\{ \left( x_{i}^{s},y_{i}^{s} \right) \right\}_{i = 1}^{n_{s}}\)
denote the labelled source dataset and
\(\mathcal{D}_{t} = \left\{ x_{j}^{t} \right\}_{j = 1}^{n_{t}}\) the
unlabelled target dataset. We assume a model \(h\) comprising a feature
extractor \(f\) and prediction head \(g\), such that \(h = g \circ f\),
with corresponding feature representations
\(z_{i}^{s} = f\left( x_{i}^{s} \right),\ \ z_{i}^{t} = f\left( x_{j}^{t} \right)\).
Since the target data are unlabelled, UDA optimises a supervised task
loss $L_{\mathrm{task}}\left( h\left( x^{s} \right),y^{s} \right)$ on the source data, together with a distribution-matching
regulariser $L_{\mathrm{disc}}\left( z^{s},z^{t} \right)$ which aligns the two feature distributions.

This paper considers two specific options for \(L_{\mathrm{disc}}\), the MMD and
CORAL. The (squared) MMD loss is defined as
\begin{equation}L_{\MMD} = \left\| \mu_{s} - \mu_{t} \right\|_{\mathcal{H}}^{2},\end{equation}
where
\(\mu_{s} = \mathbb{E}\left\lbrack \phi\left( z^{s} \right) \right\rbrack,\mu_{t} = \mathbb{E}\left\lbrack \phi\left( z^{t} \right) \right\rbrack\),
\(\mathcal{H}\) is a reproducing kernel Hilbert space, and
\(\phi\ :\mathcal{Z \rightarrow H}\) is an implicit mapping.
\(\mathcal{H}\) is associated with a unique positive-definite kernel
\(\kappa\ :\mathcal{Z \times Z}\mathbb{\rightarrow R}\) for which the
reproducing property
\(\kappa(z,z') = \left\langle \phi(z),\phi(z') \right\rangle_{\mathcal{H}}\)
is satisfied. On the other hand, CORAL aims to minimise the (squared)
Frobenius distance between the source and target feature covariance
matrices:
\begin{equation}L_{\CORAL} = \left\| \Sigma_{s} - \Sigma_{t} \right\|_{F}^{2}.\end{equation}
Define the kernel mean difference \(D_{\MMD} = \mu_{s} - \mu_{t}\) and
covariance difference \(D_{\CORAL} = \Sigma_{s} - \Sigma_{t}\). We will
also use \(D\) to refer generically to both quantities, and
\(\left\| \cdot \right\|\) for the associated norm.

In minibatch stochastic optimisation, we repeatedly sample index subsets
\(B_{s} \subseteq \left\{ 1,\ldots,n_{s} \right\}\) and
\(B_{t} \subseteq \left\{ 1,\ldots,n_{t} \right\}\), with a minibatch size of \(k\), and
compute stochastic losses \({\widehat{L}}_{{\mathrm{task}}}\left( B_{s} \right)\)
and \({\widehat{L}}_{\mathrm{disc}}\left( B_{s},B_{t} \right)\). For both the MMD
and CORAL, we use the V-statistic estimators
\({\widehat{L}}_{\mathrm{disc}} = \left\| \widehat{D} \right\|^{2}\), with
\({\widehat{D}}_{\MMD} = {\widehat{\mu}}_{s} - {\widehat{\mu}}_{t}\) and
\({\widehat{D}}_{\CORAL} = {\widehat{\Sigma}}_{s} - {\widehat{\Sigma}}_{t}\)
respectively.

We then compute stochastic gradients \(\widehat{\nabla}L_{\mathrm{disc}}\) and
\(\widehat{\nabla}L_{\mathrm{task}}\), which are noisy estimates of the
corresponding full gradients. Since the discrepancy gradients are the
main source of instability during training, our aim is to use SVR to
reduce the discrepancy gradient error
\(\mathbb{E}\left\| \widehat{\nabla}L_{\mathrm{disc}} - \nabla L_{\mathrm{disc}} \right\|^{2}\).
Since, by the chain rule,
\(\widehat{\nabla}L_{\mathrm{disc}} - \nabla L_{\mathrm{disc}} \propto \widehat{D} - D\),
we can do this by minimising \(\left\| \widehat{D} - D \right\|^{2}\)
directly (subject to some smoothness conditions discussed in \citet{Zhao2014AcceleratingSampling,Liu2020AcceleratingStrata}).

\subsection{Method overview}\label{method-overview}

PSDA constructs dependencies between sampled observations which minimise
\({\mathbb{E}\left\| \widehat{D} - D \right\|}^{2}\), conditioned on
jointly observing those examples in \(B_{s}\) and \(B_{t}\). We first
extract \(z_{i}^{s}\) and \(z_{j}^{t}\) for the full dataset, then
compute two sets of matchings.

\subsubsection{Stage 1: Source-target
matching}\label{stage-1-source-target-matching}

The first stage considers pairs of source-target indices. For each
source index \(i\) and target index \(j\), define
\begin{equation}C_{ij}^{st}= \mathbb{E}\left\lbrack \left\| \widehat{D} - D \right\|^{2}\  \middle| \ \ i \in B_{s},j \in B_{t} \right\rbrack.\end{equation}
Thus \(C_{ij}^{st}\) measures the expected error conditional on jointly
observing source example \(i\) and target example \(j\).
The law of total expectation ensures that summing these elements
(weighted by the probability of occurrence) recovers the unconditional
expectation. This fact also shows that the task of matching pairs under
this objective is a linear problem.

The specific assignment problem we solve is
\begin{gather}
\min_{U}{\sum_{i,j}^{}{U_{ij}C_{ij}^{st}}} \\
\text{subject to}  \ \ \ 0 \leq U_{ij} \leq 1,\\ 
\max\left(\left\lfloor \frac{n_{s}}{n_{t}} \right\rfloor,1\right) \leq \sum_{i}^{}U_{ij} \leq \left\lceil n_{s}/n_{t} \right\rceil,  \label{rowconst} \\
\max\left(\left\lfloor \frac{n_{t}}{n_{s}} \right\rfloor,1\right) \leq \sum_{j}^{}U_{ij} \leq \left\lceil n_{t}/n_{s} \right\rceil, \label{colconst}
\end{gather}
where \(U \in \left\{ 0,1 \right\}^{n_{s} \times n_{t}}\) is the
assignment matrix. Constraints (\ref{rowconst}) and (\ref{colconst}) deal with the imbalanced case
\(n_{s} \neq n_{t}\), and ensure that each example is assigned at least
once, while distributing the assignments as evenly as possible -- i.e.,
the number of target examples allocated to any two source examples
differs by at most one, and vice versa. Let
\(M = \left\{ \left( i_{r},j_{r} \right) \right\}_{r = 1}^{\max\left( n_{s},n_{t} \right)}\)
denote the resulting collection of matched source-target pairs, such
that \(U_{ij} = 1 \Leftrightarrow \left(i,j\right) \in M\).

\subsubsection{Stage 2: Pairing the
matches}\label{stage-2-pairing-the-matches}

The second stage of PSDA matches pairs in $M$ to form quadruplets. Thus, a
second linear assignment is solved with cost matrix
\begin{equation}C_{rq}^{MM}\mathbb{= E}\left\lbrack \left\| \widehat{D} - D \right\|^{2}\  \middle| \ \ i_{r},i_{q} \in B_{s},\ j_{r},j_{q} \in B_{t} \right\rbrack.\end{equation}

\subsubsection{Cost computation}\label{cost-computation}

In practice, computing \(C^{st}\) and \(C^{MM}\) for the MMD and CORAL
comprises computing the conditional sample kernel means and covariance
matrices respectively.

For the MMD, the source kernel mean conditioned on \(i \in B_{s}\) is
\begin{equation}{\widehat{\mu}}_{s}^{(i)} = \mathbb{E}\left\lbrack {\widehat{\mu}}_{s}\  \middle| \ i \in B_{s} \right\rbrack = \frac{\phi\left( z_{i}^{s} \right) + (k - 1)\mu_{s}}{k},\end{equation}
with an analogous expression for $\widehat{\mu}_{t}^{(j)}$. Thus, we have
\begin{equation}C_{ij}^{st} = \left\| {\widehat{\mu}}_{s}^{(i)} - {\widehat{\mu}}_{t}^{(j)} - \left( \mu_{s} - \mu_{t} \right) \right\|_{\mathcal{H}}^{2} = \frac{1}{k^{2}}\left\| \phi\left( z_{i}^{s} \right) - \phi\left( z_{j}^{t} \right) - \left( \mu_{s} - \mu_{t} \right) \right\|_{\mathcal{H}}^{2}.\end{equation}

For Stage 2, conditioning the source mean on matched pairs \(r\) and
\(q\) gives
\begin{equation}{\widehat{\mu}}_{s}^{\left( i_{r},i_{q} \right)} = \mathbb{E}\left\lbrack {\widehat{\mu}}_{s}\  \middle| \ i_{r},i_{q} \in B_{s} \right\rbrack = \frac{\phi\left( z_{i_{r}}^{s} \right) + \phi\left( z_{i_{q}}^{s} \right) + (k - 2)\mu_{s}}{k},\end{equation}
and therefore
\begin{equation}C_{rq}^{MM} = \left\| {\widehat{\mu}}_{s}^{\left( i_{r},i_{q} \right)} - {\widehat{\mu}}_{t}^{\left( j_{r},j_{q} \right)} - \left( \mu_{s} - \mu_{t} \right) \right\|_{\mathcal{H}}^{2} = \frac{1}{k^{2}}\left\| \phi\left( z_{i_{r}}^{s} \right) + \phi\left( z_{i_{q}}^{s} \right) - \phi\left( z_{j_{r}}^{t} \right) - \phi\left( z_{j_{q}}^{t} \right) - 2\left( \mu_{s} - \mu_{t} \right) \right\|_{\mathcal{H}}^{2}.\end{equation}

For the conditional sample covariance matrices, we can use the relevant
formulae for merging and updating covariance statistics \citep{Welford1962NoteProducts}, with the
appropriate finite-sample bias correction factors applied. Let
\(\widetilde{z} = z - \overline{z}\) denote centred feature vectors.
Then, for Stage 1:
\begin{equation}{\widehat{\Sigma}}_{s}^{(i)} = \mathbb{E}\left\lbrack {\widehat{\Sigma}}_{s} \middle| \ i \in B_{s} \right\rbrack = \frac{\frac{n_{s} - 1}{n_{s}}(k - 1)\Sigma_{s} + \frac{k - 1}{k}{\widetilde{z}}_{i}^{s}{{\widetilde{z}}_{i}^{s}}{}^{T}}{k - 1}.\end{equation}
For Stage 2:
\begin{equation}{\widehat{\Sigma}}_{s}^{\left( i_{r},i_{q} \right)} = \mathbb{E}\left\lbrack {\widehat{\Sigma}}_{s} \middle| \ i_{r},i_{q} \in B_{s} \right\rbrack = \frac{\frac{n_{s} - 1}{n_{s}}(k - 2)\Sigma_{s} + {\widetilde{z}}_{i_{r}}^{s}{{\widetilde{z}}_{i_{r}}^{s}}{}^{T} + {\widetilde{z}}_{i_{q}}^{s}{{\widetilde{z}}_{i_{q}}^{s}}{}^{T} - \frac{1}{k}\left( {\widetilde{z}}_{i_{r}}^{s} + {\widetilde{z}}_{i_{q}}^{s} \right)\left( {\widetilde{z}}_{i_{r}}^{s} + {\widetilde{z}}_{i_{q}}^{s} \right)^{T}}{k - 1}.\end{equation}

The costs are then obtained by substituting the corresponding covariance
matrices.

In all cases, the formulae for the cost matrices are fully vectorisable
and inexpensive to compute on a GPU. As with previous methods, the
matchings need to be periodically updated throughout training as the
feature distributions shift with each model update.

\section{Experiments}\label{experiments}

In this section, we evaluate PSDA on three criteria: degree of variance
reduction achieved, target domain accuracy, and training speed. For all
the experiments, we compare two variants of PSDA: ``Paired sampling''
performs only Stage 1 matching; ``Double-paired'' performs both Stage 1
and Stage 2.

\subsection{Estimator variance}\label{estimator-variance}

\begin{figure}
    \centering
    \includegraphics[width=0.5\linewidth]{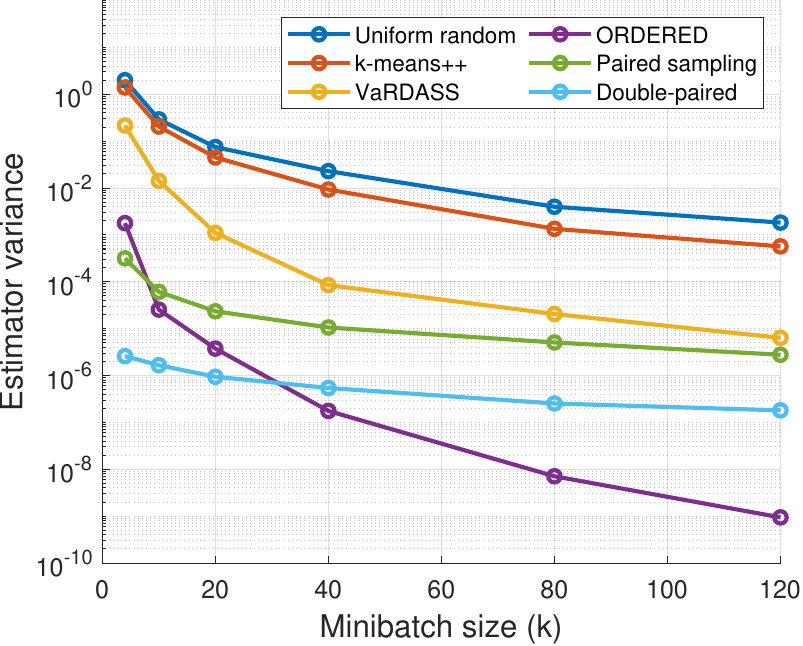}
    \caption{Estimator variance vs $k$ for six sampling algorithms.}
    \label{var}
\end{figure}

We use Monte Carlo simulations to compare the estimator variance for
different SVR methods across different values of \(k\) (Figure \ref{var}). Specifically, we
compute \({\mathbb{E\ }\left\| \widehat{D} - D \right\|}^{2}\) using a
linear kernel (that is, estimating the squared Euclidean distance
between distribution means) between a source and target dataset
comprising 2D standard normal data with \(n_{s} = n_{t} = 4,000\). In
addition to the two PSDA variants, we compare uniform random sampling,
diverse sampling using k-means++ \citep{Arthur2007K-means++:Seeding,Napoli2024ImprovingSampling}, stratified sampling (VaRDASS) \citep{Napoli2026VarianceSampling}, and
order-aware sampling (ORDERED) \citep{Napoli2026OrderData}.

For low \(k\), both PSDA variants outperform other methods, by as much
as three orders of magnitude compared to the next-best algorithm. This can be attributed to the forumulation being based on linear assignments, for which globally
optimal solutions can be found. At
the same time, the fact that observations are only considered in pairs limits the degrees of freedom of the
optimisation, resulting in a shallower curve as \(k\) increases. This
allows ORDERED to overtake PSDA for higher \(k\), although this comes at
significantly higher computational cost (see Section \ref{training speed}).

\subsection{Target domain accuracy}\label{target-domain-accuracy}

Next, we evaluate PSDA in realistic training conditions, to assess
whether the observed reduction in variance translates to an increase in
test accuracy. Experiments are conducted
using the DomainBed framework \citep{Gulrajani2021InGeneralization} on the following domain shift
benchmarks.

\textbf{Spawrious} \citep{Lynch2023Spawrious:Biases} classification of 4 dog breeds
across images with different background environments (desert, jungle,
snow etc.). This benchmark comprises 6 data splits of varying difficulty, 6 domains and 18,664 examples.

\textbf{Office-Home} \citep{Venkateswara2017DeepAdaptation} image classification of 65 categories of everyday objects with different image styles (Art, Clipart, Product, and Real World). This benchmark comprises 12 data splits, 4 domains and 15,500 examples.

\textbf{Humpbacks} \citep{Napoli2023UnsupervisedCalls} detection of humpback whale
vocalisations in underwater acoustic recordings across data from
different acoustic monitoring programs. This benchmark comprises 4 data splits, 4 domains and 8,000 examples.

The domain discrepancies are measured between the union of all training
data and a held-out subset of the evaluation set. For the MMD, we use a
radial basis function (RBF) mixture kernel \citep{Li2018DomainLearning}, given by
\(\kappa(z,z') = \sum_{\gamma \in \mathcal{G}}^{}e^{- \gamma\left\| z - z' \right\|^{2}}\)with
\(\mathcal{G} = \{ 0.001,0.01,0.1,1,10\}\).
For Spawrious and Office-Home, we use a ResNet-18 \citep{He2015DeepRecognition} backbone pre-trained on ImageNet. For
Humpbacks, we use the audio frontend and architecture described in
\citet{Napoli2023UnsupervisedCalls}. Models are trained using the Adam optimiser \citep{Kingma2014Adam:Optimization} for 3,000
iterations. Hyperparameters, including the learning rate, weight decay, minibatch size $k$, and UDA trade-off parameter, are tuned with a random search of size 10
using an in-distribution (training domain) validation set, independently
for each sampler. In particular, the random search distribution for $k$ is $k\sim2^{\text{Uniform}(3,7)}$. The entire set of experiments is repeated 5 times for
reproducibility, using different random seeds for hyperparameters,
weight initialisations, and dataset splits. All other hyperparameter
choices and training details follow the DomainBed default options.

In addition to PSDA, five variance-reduced samplers are compared: k-means++, DPP \citep{Zhang2017DeterminantalDiversification,Napoli2024ImprovingSampling}, anticlustering \citep{Baumann2026AAlgorithm}, VaRDASS, and ORDERED. We also compare several baseline UDA methods: DANN \citep{Ganin2015Domain-AdversarialNetworks}, CDAN \citep{Long2017ConditionalAdaptation}, SDAT \citep{Rangwani2022ATraining}, ELS \citep{Zhang2023FreeSmoothing}, ARM \citep{Zhang2020AdaptiveShift}, and MCC \citep{Jin2020MinimumAdaptation}, plus non-adaptive training via ERM \citep{Vapnik1998StatisticalTheory}.

Tables
\ref{spawrious split}, \ref{office split}, and \ref{humps split} show the average test accuracy and
standard errors over the 5 repeats and each of the data splits, for
each method. The results confirm the importance of
effective variance reduction when estimating UDA losses. This allows the classical MMD and CORAL techniques to outperform significantly more modern and complicated UDA methods. PSDA consistently achieves the highest or second-highest average accuracy compared to the other samplers.

\subsection{Training speed} \label{training speed}

Overall wall-clock training times are reported for each sampler and dataset in Table \ref{times}. These are averaged over all data splits, hyperparameters, and repeats from the previous experiment, including both CORAL and MMD. It can be seen that as well as being one of the highest-performing, PSDA is also among the fastest of the SVR techniques.

\section{Conclusion}\label{conclusion}

This paper introduced PSDA, a novel SVR
method for UDA based on paired sampling. We derive pairing objectives which minimise the gradient error of the MMD and CORAL losses, and demonstrate significant advantages over preceding methods in terms of computational cost and target-domain performance. Future work could include extending PSDA to other losses (both for UDA and more generally), or investigating combining PSDA with other SVR techniques.

\section{Acknowledgements}\label{acknowledgements}

The author acknowledges the use
of the IRIDIS High Performance Computing Facility, and associated
support services at the University of Southampton, in the completion of
this work.

\begin{table*}[t] \scriptsize
\centering
\caption{Average test accuracy for Spawrious by data split.}
\label{spawrious split}
\begin{tabular}{@{}l|cccccc|c@{}}
\toprule
\textbf{Method}  & \textbf{O2O-Easy}   & \textbf{O2O-Medium} & \textbf{O2O-Hard}   & \textbf{M2M-Easy}   & \textbf{M2M-Medium}  & \textbf{M2M-Hard}    & \textbf{Average}    \\ \midrule
ERM              & 68.6 ± 1.7          & 62.6 ± 0.8          & 62.1 ± 0.7          & 70.2 ± 1.8          & 45.0 ± 1.3           & 43.0 ± 1.2           & 58.6 ± 0.5          \\
DANN             & 91.4 ± 3.0          & 57.1 ± 3.5          & 71.1 ± 3.2          & 91.1 ± 0.1          & 54.8 ± 4.4           & 39.8 ± 3.4           & 67.5 ± 1.3          \\
CDAN             & 91.9 ± 1.7          & 57.0 ± 3.0          & 70.3 ± 2.2          & 92.9 ± 0.9          & 58.3 ± 3.8           & 44.3 ± 7.7           & 69.1 ± 1.6          \\
CDAN + SDAT      & 92.9 ± 1.4          & 54.3 ± 3.5          & 73.7 ± 6.6          & 83.3 ± 3.0          & 60.3 ± 4.3           & 53.0 ± 6.0           & 69.6 ± 1.8          \\
CDAN + ELS       & 89.8 ± 1.9          & 58.4 ± 2.1          & 67.3 ± 1.5          & 89.9 ± 2.2          & 62.3 ± 2.3           & 56.1 ± 10.2          & 70.6 ± 1.9          \\
ARM              & 70.2 ± 2.9          & 58.6 ± 2.1          & 60.6 ± 0.2          & 68.7 ± 1.6          & 42.2 ± 1.8           & 41.7 ± 1.0           & 57.0 ± 0.7          \\
MCC              & 87.6 ± 1.9          & 51.0 ± 0.8          & 54.0 ± 6.3          & 77.5 ± 2.4          & 46.4 ± 0.5           & 42.7 ± 1.2           & 59.9 ± 1.2          \\ \midrule
CORAL            & 70.7 ± 2.3          & 58.4 ± 1.9          & 64.1 ± 0.6          & 78.6 ± 1.5          & 54.1 ± 1.2           & 49.2 ± 0.7           & 62.5 ± 0.6          \\
+ k-means++      & 82.8 ± 3.5          & 58.2 ± 2.4          & 61.4 ± 4.1          & 75.5 ± 2.7          & 54.7 ± 2.7           & 48.6 ± 1.1           & 63.5 ± 1.2          \\
+ DPP            & 79.8 ± 3.4          & 59.8 ± 2.3          & 67.8 ± 2.0          & 79.6 ± 2.3          & 58.5 ± 1.4           & 49.4 ± 1.9           & 65.8 ± 0.9          \\
+ Anticlustering & 89.1 ± 4.7          & 57.6 ± 3.3          & 73.7 ± 6.1          & 87.5 ± 2.2          & 51.4 ± 3.2           & 47.5 ± 2.5           & 67.8 ± 1.6          \\
+ VaRDASS        & 90.1 ± 2.2          & \textbf{62.2 ± 1.0} & \textbf{79.6 ± 2.8} & 77.0 ± 2.7          & 51.7 ± 1.6           & 45.8 ± 0.4           & 67.7 ± 0.8          \\
+ ORDERED        & 88.2 ± 2.2          & 61.6 ± 1.6          & 78.1 ± 3.5          & 84.1 ± 5.0          & 60.5 ± 2.6           & 50.5 ± 2.1           & \textbf{70.5 ± 1.3} \\
+ Paired         & 87.9 ± 5.6          & 54.8 ± 1.1          & 74.4 ± 0.7          & 82.6 ± 0.9          & \textbf{61.6 ± 4.8}  & 53.9 ± 0.3           & 69.2 ± 1.3          \\
+ Double-paired  & \textbf{92.4 ± 2.2} & 53.7 ± 1.7          & 73.9 ± 7.2          & \textbf{86.0 ± 1.8} & 60.9 ± 1.7           & \textbf{54.0 ± 1.3}  & 70.2 ± 1.4          \\ \midrule
MMD              & 79.2 ± 3.3          & 61.9 ± 1.2          & 65.5 ± 3.4          & 76.2 ± 3.4          & 55.3 ± 3.4           & 48.1 ± 0.7           & 64.4 ± 1.1          \\
+ k-means++      & 83.7 ± 6.3          & 58.6 ± 2.4          & 68.4 ± 4.5          & 79.3 ± 2.7          & 60.0 ± 3.0           & 52.5 ± 4.5           & 67.1 ± 1.7          \\
+ DPP            & 83.6 ± 4.5          & \textbf{62.9 ± 0.9} & 63.5 ± 3.2          & 79.1 ± 3.1          & 57.4 ± 4.0           & 45.6 ± 1.7           & 65.4 ± 1.3          \\
+ Anticlustering & 74.5 ± 6.0          & 61.7 ± 1.0          & 85.9 ± 5.7          & 75.4 ± 4.1          & 63.7 ± 10.8          & 44.6 ± 3.0           & 67.6 ± 2.4          \\
+ VaRDASS        & \textbf{94.2 ± 1.8} & 61.5 ± 1.4          & 72.7 ± 4.5          & 76.9 ± 4.3          & \textbf{75.9 ± 10.6} & 48.1 ± 5.1           & 71.6 ± 2.2          \\
+ ORDERED        & 93.5 ± 1.3          & 56.4 ± 2.4          & \textbf{85.1 ± 1.9} & \textbf{88.6 ± 0.8} & 70.5 ± 7.8           & 62.1 ± 10.9          & \textbf{76.1 ± 2.3} \\
+ Paired         & 86.1 ± 6.1          & 59.8 ± 5.4          & 85.2 ± 3.5          & 78.1 ± 6.4          & 68.2 ± 8.2           & 55.6 ± 12.0          & 72.2 ± 3.0          \\
+ Double-paired  & 87.2 ± 6.7          & 62.4 ± 5.5          & 77.4 ± 8.9          & 86.6 ± 5.5          & 60.9 ± 11.2          & \textbf{67.5 ± 13.0} & 73.7 ± 3.6          \\ \bottomrule
\end{tabular}
\end{table*}

\begin{table*}[t] \scriptsize
\caption{Average test accuracy for Office-Home by data split.}
\label{office split}
\centering
\begin{tabular}{@{}l|cccccccccccc|c@{}}
\toprule
\textbf{Method}  & \textbf{C-A}  & \textbf{A-C}  & \textbf{P-A}  & \textbf{A-P}  & \textbf{R-A}  & \textbf{A-R}  & \textbf{P-C}  & \textbf{C-P}  & \textbf{R-C}  & \textbf{C-R}  & \textbf{R-P}  & \textbf{P-R}  & \textbf{Average}    \\ \midrule
ERM              & 32.9          & 33.9          & 30.7          & 45.7          & 49.1          & 57.5          & 34.2          & 46.2          & 36.5          & 51.8          & 63.2          & 59.6          & 45.1 ± 0.4          \\
DANN             & 31.1          & 34.9          & 28.9          & 39.9          & 46.9          & 50.5          & 31.7          & 46.8          & 39.7          & 48.7          & 63.4          & 52.7          & 42.9 ± 0.4          \\
CDAN             & 36.8          & 31.2          & 30.0          & 40.5          & 47.3          & 52.9          & 33.3          & 44.4          & 42.6          & 49.3          & 60.7          & 54.8          & 43.7 ± 0.5          \\
CDAN + SDAT      & 36.0          & 36.1          & 30.9          & 41.9          & 48.0          & 54.5          & 37.8          & 45.7          & 43.5          & 50.5          & 66.9          & 55.7          & 45.6 ± 0.2          \\
CDAN + ELS       & 35.1          & 30.7          & 29.3          & 38.3          & 45.7          & 52.9          & 33.6          & 44.3          & 41.6          & 46.7          & 62.7          & 55.2          & 43.0 ± 0.3          \\
ARM              & 34.2          & 31.3          & 30.0          & 43.1          & 49.2          & 56.5          & 32.6          & 46.4          & 35.6          & 47.6          & 63.1          & 56.0          & 43.8 ± 0.2          \\
MCC              & 33.8          & 38.3          & 34.3          & 50.7          & 49.9          & 58.7          & 36.4          & 54.1          & 44.2          & 55.6          & 69.1          & 59.8          & 48.7 ± 0.3          \\ \midrule
CORAL            & 39.4          & 35.2          & 34.7          & 40.8          & 57.1          & 56.8          & 35.8          & 48.3          & 43.2          & 49.5          & 68.9          & 60.0          & 47.5 ± 0.1          \\
+ k-means++      & 41.5          & \textbf{40.3} & 40.0          & 46.1          & 53.3          & 56.7          & 40.1          & 51.0          & 45.6          & 55.7          & 70.6          & 64.1          & 50.4 ± 0.5          \\
+ DPP            & 40.8          & 36.0          & 36.9          & 40.5          & 54.5          & 56.3          & 35.8          & 48.7          & 43.5          & 49.6          & 68.9          & 61.0          & 47.7 ± 0.2          \\
+ Anticlustering & 39.2          & 37.2          & 36.8          & 40.4          & 56.5          & 56.8          & 38.4          & 49.9          & 45.4          & 54.2          & 70.1          & 62.6          & 49.0 ± 0.2          \\
+ VaRDASS        & 39.7          & 35.6          & 36.6          & 42.8          & 55.7          & 57.9          & 39.3          & 50.3          & 47.3          & 52.8          & 71.6          & 62.3          & 49.3 ± 0.4          \\
+ ORDERED        & 41.9          & 37.3          & 39.6          & 44.9          & 58.6          & 58.3          & 42.2          & 51.4          & 47.4          & 55.8          & 71.5          & 64.0          & 51.1 ± 0.4          \\
+ Paired         & 44.1          & 39.9          & 40.7          & \textbf{46.2} & 57.9          & 59.6          & 43.6          & 52.9          & \textbf{51.1} & \textbf{57.2} & 71.2          & \textbf{65.9} & 52.5 ± 0.2          \\
+ Double-paired  & \textbf{44.2} & \textbf{40.3} & \textbf{41.2} & \textbf{46.2} & \textbf{58.7} & \textbf{60.0} & \textbf{44.5} & \textbf{54.0} & 50.6          & 55.9          & \textbf{71.9} & 65.1          & \textbf{52.7 ± 0.4} \\ \midrule
MMD              & 32.4          & 35.4          & 31.5          & 47.0          & 49.5          & 55.7          & 32.2          & 48.4          & \textbf{41.7} & 49.3          & 66.6          & 56.0          & 45.5 ± 0.3          \\
+ k-means++      & 33.6          & 33.7          & 32.4          & 43.9          & \textbf{51.8} & 53.4          & 31.9          & 48.0          & 39.3          & \textbf{52.5} & 66.2          & 55.5          & 45.2 ± 0.2          \\
+ DPP            & 31.6          & 34.9          & 31.0          & 45.2          & 51.0          & 56.3          & 33.4          & \textbf{51.2} & 39.2          & 50.1          & \textbf{67.2} & 59.0          & 45.9 ± 0.4          \\
+ Anticlustering & 35.1          & 33.2          & 33.0          & 44.6          & 50.1          & 57.2          & 33.3          & 48.4          & 36.6          & 51.0          & 63.8          & 58.3          & 45.4 ± 0.5          \\
+ VaRDASS        & 33.8          & 33.5          & 32.4          & 45.8          & 50.4          & 57.0          & 32.5          & 49.8          & 37.7          & 51.8          & 65.3          & 58.8          & 45.7 ± 0.2          \\
+ ORDERED        & \textbf{35.9} & \textbf{36.4} & \textbf{33.2} & \textbf{47.1} & 48.7          & 55.7          & 32.7          & 49.8          & 39.0          & 52.3          & 66.2          & 59.8          & \textbf{46.4 ± 0.2} \\
+ Paired         & 35.1          & 33.8          & 29.6          & 43.9          & 49.6          & 55.1          & \textbf{34.5} & 50.4          & 37.5          & 51.8          & 64.4          & 58.8          & 45.4 ± 0.2          \\
+ Double-paired  & 33.5          & 35.1          & 31.6          & 45.8          & \textbf{51.8} & \textbf{57.3} & 33.4          & 49.1          & 38.5          & 52.0          & 65.4          & \textbf{61.5} & 46.3 ± 0.3          \\ \bottomrule
\end{tabular}
\end{table*}

\begin{table*}[t] \scriptsize
\centering
\caption{Average test accuracy for Humpbacks by data split.}
\label{humps split}
\begin{tabular}{@{}l|cccc|c@{}}
\toprule
\textbf{Method}  & \textbf{Domain 1}   & \textbf{Domain 2}   & \textbf{Domain 3}   & \textbf{Domain 4}   & \textbf{Average}    \\ \midrule
ERM              & 70.3 ± 2.7          & 92.0 ± 1.9          & 78.1 ± 3.0          & 96.2 ± 0.6          & 84.2 ± 1.1          \\
DANN             & 60.5 ± 4.2          & 90.1 ± 2.1          & 63.9 ± 6.1          & 76.1 ± 11.1         & 72.6 ± 3.4          \\
CDAN             & 61.6 ± 4.2          & 82.2 ± 4.7          & 73.5 ± 3.0          & 84.4 ± 0.6          & 75.4 ± 1.7          \\
CDAN + SDAT      & 63.7 ± 3.0          & 81.2 ± 6.7          & 63.6 ± 3.7          & 78.9 ± 3.2          & 71.8 ± 2.2          \\
CDAN + ELS       & 62.7 ± 2.4          & 85.6 ± 3.4          & 70.8 ± 2.2          & 83.9 ± 2.3          & 75.8 ± 1.3          \\
ARM              & 78.8 ± 3.4          & 95.9 ± 1.0          & 72.4 ± 3.6          & 89.6 ± 2.0          & 84.2 ± 1.4          \\
MCC              & 70.0 ± 5.9          & 87.7 ± 4.9          & 76.1 ± 4.1          & 97.1 ± 0.5          & 82.7 ± 2.2          \\ \midrule
CORAL            & 77.2 ± 1.4          & 87.7 ± 3.4          & 83.6 ± 4.2          & 92.7 ± 1.9          & 85.3 ± 1.5          \\
+ k-means++      & 79.1 ± 1.7          & 97.8 ± 0.6          & 85.4 ± 4.6          & 93.8 ± 1.0          & 89.1 ± 1.3          \\
+ DPP            & \textbf{81.0 ± 1.9} & 97.4 ± 1.0          & 84.4 ± 5.1          & 94.9 ± 1.2          & 89.5 ± 1.4          \\
+ Anticlustering & 78.8 ± 1.6          & 95.6 ± 2.0          & 87.0 ± 5.9          & 93.1 ± 2.5          & 88.6 ± 1.7          \\
+ VaRDASS        & 78.4 ± 2.8          & 95.1 ± 1.6          & 85.3 ± 4.2          & 94.4 ± 1.2          & 88.3 ± 1.3          \\
+ ORDERED        & 79.8 ± 1.9          & \textbf{98.0 ± 0.9} & \textbf{92.0 ± 3.1} & 93.8 ± 1.8          & 90.9 ± 1.0          \\
+ Paired         & 80.2 ± 2.3          & 94.1 ± 1.5          & 90.5 ± 3.2          & 95.8 ± 0.7          & 90.1 ± 1.1          \\
+ Double-paired  & 80.7 ± 1.4          & 97.5 ± 0.9          & 89.9 ± 2.9          & \textbf{96.0 ± 0.7} & \textbf{91.0 ± 0.9} \\ \midrule
MMD              & 78.3 ± 2.5          & 95.0 ± 1.2          & 83.3 ± 4.2          & 94.3 ± 1.1          & 87.7 ± 1.3          \\
+ k-means++      & 79.7 ± 1.9          & 97.6 ± 0.4          & 79.5 ± 5.7          & \textbf{96.7 ± 0.3} & 88.4 ± 1.5          \\
+ DPP            & 83.4 ± 1.0          & 97.9 ± 0.6          & 83.4 ± 5.9          & 96.6 ± 0.6          & 90.3 ± 1.5          \\
+ Anticlustering & 80.5 ± 2.4          & 95.2 ± 0.9          & 92.3 ± 4.7          & 93.8 ± 0.5          & 90.5 ± 1.3          \\
+ VaRDASS        & 79.2 ± 1.1          & \textbf{98.4 ± 0.5} & \textbf{95.5 ± 1.2} & 95.0 ± 1.0          & 92.0 ± 0.5          \\
+ ORDERED        & 78.9 ± 2.0          & 98.1 ± 0.3          & 94.8 ± 2.6          & 95.5 ± 1.0          & 91.8 ± 0.9          \\
+ Paired         & \textbf{81.5 ± 1.4} & 95.6 ± 0.9          & 91.9 ± 2.4          & 94.6 ± 0.4          & 90.9 ± 0.7          \\
+ Double-paired  & 80.2 ± 2.3          & 97.3 ± 0.7          & 95.3 ± 0.9          & 95.6 ± 0.6          & \textbf{92.1 ± 0.7} \\ \bottomrule
\end{tabular}
\end{table*}

\begin{table}[t]
\caption{Average wall-clock training times by dataset (seconds).}
\label{times}
\centering
\begin{tabular}{@{}l|ccc@{}}
\toprule
\multicolumn{1}{c|}{\textbf{Sampler}}  & \textbf{Spawrious} & \textbf{Office-Home} & \textbf{Humpbacks} \\ \midrule
Uniform random                         & 393 ± 9            & 501 ± 18             & 30 ± 1             \\
k-means++                              & 2312 ± 151         & 997 ± 28             & 345 ± 15           \\
DPP                                    & 4123 ± 109         & 1095 ± 32            & 731 ± 19           \\
Anticlustering                         & 1403 ± 24          & 818 ± 25             & 124 ± 2            \\
VaRDASS                                & 2958 ± 21          & 1166 ± 45            & 528 ± 9            \\
ORDERED                                & 3900 ± 83          & 2579 ± 97            & 1944 ± 43          \\
Paired                                 & 1218 ± 14          & 586 ± 18             & 232 ± 2            \\
Double-paired                          & 1329 ± 13          & 604 ± 20             & 286 ± 3            \\ \bottomrule
\end{tabular}
\end{table}

{
    \small
    \bibliographystyle{tmlr}
    \bibliography{references}
}

\end{document}